\def\year{2018}\relax
\documentclass[letterpaper]{article}
\usepackage{aaai18}
\usepackage{times}
\usepackage{lipsum}
\usepackage{helvet}
\usepackage{courier}
\usepackage{mathtools}
\usepackage{amsmath}
\usepackage{latexsym}
\usepackage{float}
\usepackage{graphicx}
\usepackage{algorithm}
\usepackage{algpseudocode}
\usepackage{amssymb}
\usepackage{array}
\usepackage{multirow}
\usepackage{booktabs}
\usepackage{siunitx}
\usepackage{url}
\usepackage{tablefootnote}
\usepackage{amsthm}
\usepackage{amssymb}
\usepackage{pifont}
\usepackage{array}
\usepackage{xcolor}

\usepackage{soul}

\renewcommand{\footnotesize}{\small}
\usepackage[hang,flushmargin]{footmisc}

\usepackage[]{caption}

\definecolor{8red}{HTML}{A70000}
\definecolor{6red}{HTML}{FF0000}
\definecolor{4red}{HTML}{FF5252}
\definecolor{2red}{HTML}{FF7B7B}
\definecolor{1red}{HTML}{FFBABA}

\newcolumntype{x}[1]{>{\centering\arraybackslash\hspace{0pt}}p{#1}}

\title{\textsc{SkipFlow:} Incorporating Neural Coherence Features \\ for End-to-End Automatic Text Scoring}

\author{Yi Tay\textsuperscript{$1$}, Minh C. Phan\textsuperscript{$2$}, Luu Anh Tuan\textsuperscript{$3$} \and Siu Cheung Hui\textsuperscript{$4$}\\
\textsuperscript{$1,2,4$}\:Nanyang Technological University \\ School of Computer Science and Engineering, Singapore \\
\textsuperscript{$3$}\:Institute for Infocomm Research, Singapore \\
}

\date{}

\begin{document}
\maketitle
\begin{abstract}
Deep learning has demonstrated tremendous potential for Automatic Text Scoring (ATS) tasks. In this paper, we describe a new neural architecture that enhances vanilla neural network models with auxiliary neural coherence features. Our new method proposes a new \textsc{SkipFlow} mechanism that models relationships between snapshots of the hidden representations of a long short-term memory (LSTM) network as it reads. Subsequently, the semantic relationships between multiple snapshots are used as auxiliary features for prediction. This has two main benefits. Firstly, essays are typically long sequences and therefore the memorization capability of the LSTM network may be insufficient. Implicit access to multiple snapshots can alleviate this problem by acting as a protection against vanishing gradients. The parameters of the \textsc{SkipFlow} mechanism also acts as an auxiliary memory. Secondly, modeling relationships between multiple positions allows our model to learn features that represent and approximate textual coherence. In our model, we call this \textit{neural coherence} features. Overall, we present a unified deep learning architecture that generates neural coherence features as it reads in an end-to-end fashion. Our approach demonstrates state-of-the-art performance on the benchmark ASAP dataset, outperforming not only feature engineering baselines but also other deep learning models. 
\end{abstract}

\section{Introduction}

Automated Text Scoring (ATS) systems are targeted at both alleviating the workload of teachers and improving the feedback cycle in educational systems. ATS systems have also seen adoption for several high-stakes assessment, e.g., the \textit{e-rater system} \cite{attali2004automated} which has been used for TOEFL and GRE examinations. A successful ATS system brings about widespread benefits to society and the education industry. This paper presents a novel neural network architecture for this task. 

Traditionally, the task of ATS has been regarded as a machine learning problem \cite{larkey1998automatic,attali2004automated} which learns to approximate the marking process with supervised learning. Decades of ATS research follow the same traditional supervised text regression methods in which handcrafted features are constructed and subsequently passed into a machine learning based classifier. A wide assortment of features may commonly extracted from essays. Simple and intuitive features may include essay length, sentence length. On the other hand, intricate and complex features may also be extracted, e.g.., features such as grammar correctness \cite{attali2004automated}, readability \cite{DBLP:conf/bea/ZeschWS15} and textual coherence \cite{DBLP:conf/emnlp/ChenH13}. However, these handcrafted features are often painstakingly designed, require a lot of human involvement and usually require laborious implementation for every new feature.

Deep learning based ATS systems have recently been proposed \cite{DBLP:conf/emnlp/DongZ16,DBLP:conf/emnlp/TaghipourN16,DBLP:conf/acl/AlikaniotisYR16}. A comprehensive study has been done in \cite{DBLP:conf/emnlp/TaghipourN16} which demonstrated that neural network architectures such as the long short-term memory (LSTM) \cite{hochreiter1997long} and convolutional neural network (CNN) are capable of outperforming systems that extensively require handcrafted features. However, all of these neural models do not consider transition of an essay over time, i.e., logical flow and coherence over time. In particular, mainly semantic compositionality is modeled within the recursive operations in the LSTM model which compresses the input text repeatedly within the recurrent cell. In this case, the relationships between multiple points in the essay cannot be captured effectively. Moreover, essays are typically long sequences which pushes the limits of the memorization capability of the LSTM.

Hence, the objective of this work is a unified solution to the above mentioned problems. Our method alleviates two problems. The first is targeted at alleviating the inability of current neural network architectures to model flow, coherence and semantic relatedness over time. The second is aimed at easing the burden of the recurrent model. In order to do so, we model the relationships between multiple snapshots of the LSTM's hidden state over time. More specifically, as our model reads the essay, it models the semantic relationships between two points of an essay using a neural tensor layer. Eventually, \textit{multiple} features of semantic relatedness are aggregated across the essay and used as auxiliary features for prediction. 

The intuition behind our idea is as follows. Firstly, semantic relationships across sentences are commonly used as an indicator of writing flow and textual coherence \cite{wiemer2000select,higgins2004evaluating,higgins2007sentence,DBLP:conf/emnlp/ChenH13,somasundaran2014lexical}. As such, our auxiliary features (generated end-to-end) aim to capture the logical and semantic flow of an essay. This also provides a measure of \textit{semantic similarity} aside from the flavor of \textit{semantic compositionality} modeled by the base LSTM model. 

Secondly, the additional parameters from the external tensor serve as an auxiliary memory for the network. As essays are typically long sequences, modeling the relationship between distant states with additional parameters can enhance memorization and improve performance of the deep architecture by allowing access to intermediate states, albeit implicitly. The semantic relevance scores can then be aggregated by concatenation and passed as an auxiliary feature to a fully-connected dense layer in the final layer of the network. As such, our architecture performs sentence modeling (compositional reading) and semantic matching in a unified end-to-end framework. 

\subsection{Our Contributions}

The prime contributions of our paper are as follows:
\begin{itemize}
\item For the first time, we consider neural coherence features within the context of an end-to-end neural framework. Semantic similarity and textual coherence have a long standing history in ATS literature \cite{wiemer2000select,higgins2007sentence,higgins2004evaluating}. Our work incorporates this intuition into modern neural architectures. 
\item Aside from modeling coherence, our method also alleviates and eases the burden of the recurrent model by implicit access to hidden representations over time. This serves as a protection against vanishing gradient. Moreover, a better performance can be achieved with a smaller LSTM parameterization. 
\item We propose \textsc{SkipFlow} LSTM, a new neural architecture that incorporates the intuition of logical and semantic flow into the vanilla LSTM model. \textsc{SkipFlow} LSTM obtains state-of-the-art performance on the ASAP benchmark dataset. We also achieve an increase of $6\%$ in performance over a strong feature engineering baseline. In the same experimental configuration, we achieve about $10\%$ increase over a baseline LSTM model, outperforming more advanced extensions such as Multi-Layered LSTMs and attention-based LSTMs. 
\end{itemize}

\section{Related Work}
Automated Text Scoring (ATS) systems have been deployed for high-stakes assessment since decades ago. Early high-stakes ATS systems include the Intelligent Essay Assessor (IEA) \cite{foltz2013implementation} and Project Essay Grade \cite{page1967grading,shermis2003automated}. Commercial ATS systems such as the \textit{e-rater} \cite{attali2004automated} have been also deployed for GRE and TOEFL examinations. 

Across the rich history of ATS research, supervised learning based ATS systems mainly rely on domain-specific feature engineering whereby lexical, syntactic and semantic features are designed by domain experts and subsequently extracted from essays. Then, a simple machine learning classifier trained on these feature vectors can be used to predict the grades of essays. Early work \cite{larkey1998automatic} treats ATS as a text categorization problem and uses a Naive Bayes model for grading while the \textit{e-rater} system uses linear regression over handcrafted features. \cite{DBLP:conf/emnlp/PhandiCN15} proposed a Bayesian Linear Ridge Regression approach for domain adaptation of essays. 

 The reliance on handcrafted features is a central theme to decades of ATS research. The complexity and ease of implementation of essay scoring features can be diverse. For example, length-based features are intuitive and simple to extract from essays. On the other hand, there are more complex features such as grammar correctness or lexical complexity. Features such as readability \cite{DBLP:conf/bea/ZeschWS15}, textual and discourse coherence \cite{DBLP:conf/emnlp/ChenH13,somasundaran2014lexical} are also harder to design in which convoluted pipelines have to be built for feature extraction to be performed. As a whole, feature engineering is generally a laborious process, i.e., apart from designing features, custom code has to be written for each additional feature. For a comprehensive review of feature engineering in the context of ATS, we refer interested readers to \cite{DBLP:conf/bea/ZeschWS15}. 

Recently, motivated by the success of deep learning in many domains, several deep learning architectures for ATS have been proposed. \cite{DBLP:conf/emnlp/TaghipourN16,DBLP:conf/emnlp/DongZ16} empirically evaluated the performance of a myriad of deep learning models on the ATS tasks. In their work, models such as the recurrent neural network (RNN) and convolutional neural network (CNN) demonstrated highly competitive results without requiring any feature engineering. On the other hand, an adapted task-specific embedding approach was proposed in \cite{DBLP:conf/acl/AlikaniotisYR16} that learns semantic word embeddings while predicting essay grades. Subsequently, these adapted word embeddings are passed as input to a LSTM network for prediction. The attractiveness of neural text scoring stems from the fact that features are learned end-to-end, diminishing the need for laborious feature engineering to be performed. 


Our work extends the vanilla model and enhances with the incorporation of \textit{neural coherence} features. The concept of semantic similarity between sentences has been used to measure coherence in student essays \cite{higgins2007sentence,higgins2004evaluating}. Textual coherence features have also been adopted in \cite{DBLP:conf/emnlp/ChenH13} which measures the semantic similarity between nouns and proper nouns. Lexical chaining \cite{somasundaran2014lexical} has also been used for measuring discourse quality in student essays. Our work, however, is the first \textit{neural coherence} model that incorporates these features into an end-to-end fashion. Different from traditional coherence features, our neural features form a part of an overall unified framework.

Our proposed approach is inspired by the field of semantic matching. In semantic matching, a similarity score is produced between two vectors and is often used in many NLP and IR applications. The usage of tensor layers and bilinear similarity is inspired by many of these works. For example, convolutional neural tensor network (CNTN) \cite{DBLP:conf/ijcai/QiuH15} and NTN-LSTM \cite{DBLP:conf/sigir/TayPLH17} are recently proposed architectures for question-answer pair matching. However, unlike ours, these works are mainly concerned with matching between two sentences and are often trained with two networks. The tensor layer, also known as the Neural Tensor Network (NTN), was first incepted as a compositional operator in Recursive Neural Networks for sentiment analysis \cite{socher2013recursive}. Subsequently, it has also been adopted for rich and expressive knowledge base completion \cite{DBLP:conf/nips/SocherCMN13}. It has also seen adoption in end-to-end memory networks \cite{DBLP:conf/cikm/TayTH17}. The NTN is parameterized by both a tensor and an ordinary linear layer in which the tensor parameters model multiple instances of second order interactions between two vectors. The adoption of the tensor layer in our framework is motivated by the strong empirical performance of NTN.

In our approach, we generate neural coherence features by performing semantic matching $k$ times while reading. This can be interpreted as jointly matching and reading. These additional parameters can also be interpreted as an auxiliary memory which can also help and ease the burden of the LSTM memory. LSTMs are known to have difficulty in modeling long term dependencies\footnote{Essays are typically long documents spanning 300-800 words on average.} and due to their compositional nature, measuring relatedness and coherence between two points becomes almost impossible. Moreover, our \textsc{SkipFlow} mechanism serves as an additional protection against the vanishing gradient problem by exposing hidden states to deeper layers. In a similar spirit, attention mechanisms \cite{bahdanau2014neural} learn a weighted combination of hidden states across all time steps and produces a global feature vector. However, our approach learns auxiliary features that are used for prediction.

\section{Our \textsc{SkipFlow} LSTM Model}
\begin{figure*}[ht]
  
  \centering
    \includegraphics[width=0.8\textwidth]{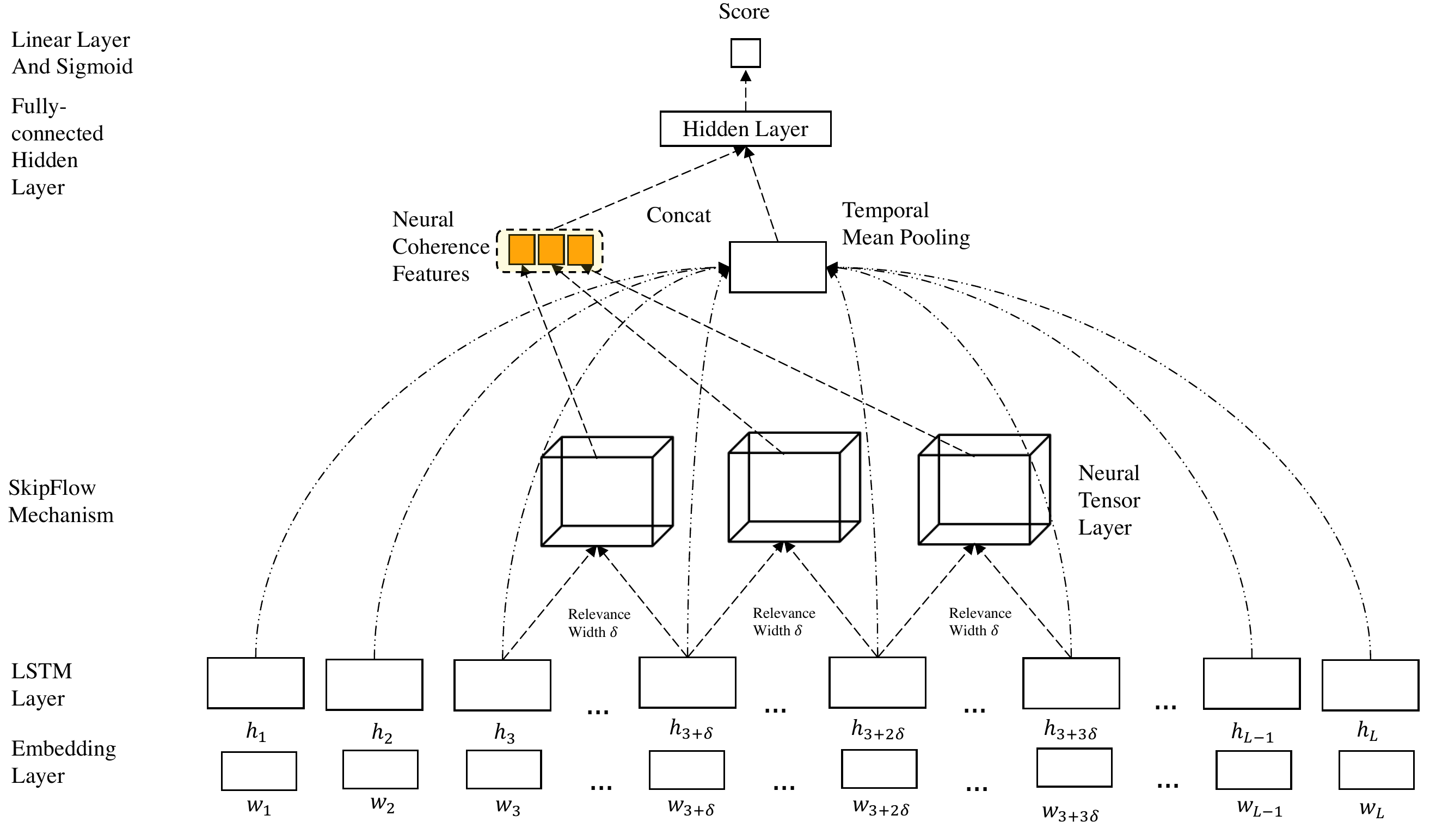}
    \caption{Illustration of our proposed \textsc{SkipFlow} LSTM model with width $\delta$. Note that tensors depicted are \textbf{shared} parameters and there is only one tensor parameter in the entire architecture.}
    \label{fig:overall}
\end{figure*}

In this section, we introduce the overall model architecture of \textsc{SkipFlow}. Figure \ref{fig:overall} depicts the proposed architecture of our model. 

\subsection{Embedding Layer}
Our model accepts an essay and the target score as a training instance. Each essay is represented as a fixed-length sequence in which we pad all sequences to the maximum length. Let $L$ be the maximum essay length. Subsequently, each sequence is converted into a sequence of low-dimensional vectors via the embedding layer. The parameters of the embedding layer are defined as $W_e \in \mathbb{R}^{|V| \times N}$ where $|V|$ is the size of the vocabulary and $N$ is the dimensionality of the word embeddings. 

\subsection{Long Short-Term Memory (LSTM)}
The sequence of word embeddings obtained from the embedding layer is then passed into a long short-term memory (LSTM) network \cite{hochreiter1997long}.

\begin{equation}
h_{t} = LSTM(h_{t-1}, x_i)
\end{equation}
where $x_t$ and $h_{t-1}$ are the input vectors at time $t$. The LSTM model is parameterized by output, input and forget gates, controlling the information flow within the recursive operation. For the sake of brevity, we omit the technical details of LSTM which can be found in many related works. At every time step $t$, LSTM outputs a hidden vector $h_t$ that reflects the semantic representation of the essay at position $t$. To select the final representation of the essay, a temporal mean pool is applied to all LSTM outputs.


\subsection{\textsc{SkipFlow} Mechanism for Generating Neural Coherence Features}
In this section, we describe the process of generating neural coherence features within our end-to-end framework. 
\subsubsection{Skipping and Relevance Width}
In our proposed approach, the relationships between two positional outputs of LSTM across time steps are modeled via a parameterized compositional technique that generates a \textit{coherence feature}. Let $\delta$ be a hyperparameter that controls the relevance width of the model. For each LSTM output, we select pairs of sequential outputs of width $\delta$, i.e., $\{ (h_i,h_{i+\delta}),(h_{i+\delta},h_{i+2\delta}),(h_{i+2X},h_{i+3\delta}),..\}$ are the tuples from the outputs that are being composed, $h_t$ denotes the output of LSTM at time step $t$. In our experiments, the starting position\footnote{We set the starting position $i>0$ to avoid matching against the initial state.} is fixed at $i=3$. For the sake of simplicity, if the width $\delta$ exceeds the max length, we loop back to the beginning of the essay in a circular fashion. The rationale for fixed length matching is as follows. Firstly, we want to limit the amount of preprocessing required as determining important key points such as nouns and pronouns require preprocessing of some sort. Secondly, maintaining specific indices for each essay can be cumbersome in the context of batch-wise training of deep learning models using libraries restricted by static computational graphs. Finally, LSTMs are memory-enabled models and therefore, intuitively, a slight degree of positional impreciseness should be tolerable. 

\subsubsection{Neural Tensor Layer}
We adopt a tensor layer to model the relationship between two LSTM outputs. The tensor layer is a parameterized composition defined as follows:
\begin{equation}
s_i(a,b) = \sigma(u^T f(v^{T}_{a} M^{[1:k]}v_b + V[v_a, v_b] +b))
\end{equation} 
where $f$ is a non-linear function such as $tanh$. $M^{[1:k]} \in \mathbb{R}^{n \times n \times k}$ is a tensor (3d matrix). $a,b \in \mathbb{R}^{d}$ are the vector outputs of LSTM at two time steps of $\delta$-width apart where $d$ is the dimensionality of LSTM parameters. For each slice of the tensor $M$, each bilinear tensor product $v^{T}_{a} M_k v_b$ returns a scalar to form a $k$ dimensional vector. $\sigma$ is the sigmoid function which constraints the output to $[0,1]$. The other parameters are the standard form of a neural network. In our model, two vectors (outputs of LSTM) are passed through the tensor layer and returns a similarity score $s_i(h_i,h_{i+X})\in[0,1]$ that determines the coherence feature between the two vectors. The parameters of the tensor layer are shared throughout all output pairs. The usage of bilinear product enables dyadic interaction between vectors through a similarity matrix. This enables a rich interaction between hidden representations. Moreover, the usage of multiple slices encourages different aspects of this relation to be modeled. 

\subsection{Fully-connected Hidden Layer}
Subsequently, all the scalar values $s_1,s_2,\cdots,s_n$ that are obtained from the tensor layer are concatenated together to form the neural coherence feature vector. $n$ is the number of times that coherence is being measured, depending on the relevance width $\delta$ and maximum sequence length $L$. Recall that the essay representation is obtained from a mean pooling over all hidden states. This essay vector is then concatenated with the coherence feature vector. This vector is then passed through a fully connected hidden layer defined as follows: 
\begin{equation}
h_{out} = f (\textbf{W}_h([e, s_1,s_2,....,s_n])) + b_h
\end{equation}
where $f(.)$ is a non-linear activation such as \textit{tanh} or \textit{relu}, $\textbf{W}_h$ and $b_h$ are the parameters of the hidden layer. $e$ is the final essay representation obtained from temporal mean pooling and $s_1,s_2,...,s_n$ are the scalar values obtained from the neural tensor layer, i.e., each scalar value is the matching score from  $\{ (h_i,h_{i+\delta}),(h_{i+\delta},h_{i+2\delta}),(h_{i+2\delta},h_{i+3\delta}),..\}$. 

\subsection{Linear Layer with Sigmoid}
Finally, we pass $h_{out}$ into a final linear regression layer. The final layer is defined as follows:
\begin{equation}
y_{out} = \sigma \: (\textbf{W}_f([h_{out}])) + b_f
\end{equation}
where $\textbf{W}_f, b_f$ are parameters of the final linear layer, $\sigma$ is the sigmoid function and $y_{out} \in [0,1]$. The output at this final layer is the normalized score of the essay. Following \cite{DBLP:conf/emnlp/TaghipourN16}, the bias is set to the mean expected score. 
 
\subsection{Learning and Optimization}
Our network optimizes the mean-square error which is defined as:

\begin{equation}
MSE(z,z^{*}) = \frac{1}{N} \sum^{N}_{i=1} (z_i - z^*_i)^2
\end{equation}

\noindent where $z_{i}^{*}$ is the gold standard score and $z_i$ is the model output. The parameters of the network are then optimized using gradient descent.

\section{Experimental Evaluation}
In this section, we describe our experimental procedure, dataset and empirical results. 
\subsection{Dataset}
We use the ASAP (Automated Student Assessment Prize) dataset for experimental evaluation. This comes from the competition which was organized and sponsored by the William and Flora Hewlett Foundation (Hewlett) and ran on Kaggle from 10/2/12 to 30/4/12. This dataset contains 8 essay prompts as described in Table \ref{tab:asap}. Each prompt can be interpreted as a different essay topic along with a different genre such as argumentative or narrative. 
 \begin{table}[htbp]
   \centering
\small
     \begin{tabular}{|c|c|c|c|}
     \hline
     Prompt & \#Essays & Avg Length & Scores \\
     \hline
     1     & 1783  & 350   & 2-12 \\
     2     & 1800  & 350   &  1-6 \\
     3     & 1726  & 150   &  0-3 \\
     4     & 1772  & 150   &  0-3 \\
     5     & 1805  & 150   &  0-4\\
     6     & 1800  & 150   &  0-4\\
     7     & 1569  & 250   &  0-30\\
     8     & 723   & 650   &  0-60\\
     \hline
     \end{tabular}%
       \caption{Statistics of ASAP dataset. Scores denote the range of possible marks in the dataset. }
   \label{tab:asap}%
 \end{table}%

\subsection{Experimental Setup}
 We use 5-fold cross validation to evaluate all systems with a 60/20/20 split for train, development and test sets. The splits are provided by \cite{DBLP:conf/emnlp/TaghipourN16} and the experimental procedure is followed closely. We train all models for 50 epochs and select the best model based on the performance on the development set. The vocabulary is restricted to the $4000$ most frequent words.  We tokenize and lowercase text using NLTK\footnote{http://www.nltk.org}, and normalize all score range to within [0,1]. The scores are rescaled back to the original prompt-specific scale for calculating Quadratic Weighted Kappa (QWK) scores. Following \cite{DBLP:conf/emnlp/TaghipourN16}, the evaluation is conducted in prompt-specific fashion. Even though training prompts together might seem ideal, it is good to note that each prompt can contain genres of essays that are very contrastive such as narrative or argumentative essays. Additionally, prompts can have different marking schemes and level of students. As such, it would be extremely difficult to train prompts together. 

\subsection{Evaluation Metric}
 The evaluation metric used is the Quadratic Weighted Kappa (QWK) which measures agreement between raters and is a commonly used metric for ATS systems. The QWK score ranges from 0 to 1 but becomes negative if there is less agreement than expected by chance. The QWK score is calculated as follows. First, an $N \times N$ histogram matrix $\textbf{O}$ is constructed. Next, a weight matrix $\textbf{W}_{i,j} = \frac{(i-j)^2}{(N-1)^2}$ is calculated that corresponds to the difference between rater's scores where $i$ and $j$ are reference ratings by the annotator and the ATS system. Finally, another $N\times N$ histogram matrix $\textbf{E}$ is constructed assuming no correlation between rating scores. This is done using an outer product between each rater's histogram vector and normalized such that $sum(\textbf{E}) = sum(\textbf{O})$. Finally, the QWK score is calculated as $\kappa = 1 - \frac{\sum_{i,j}w_{i,j}O_{i,j}}{\sum_{i,j}w_{i,j}E_{i,j}}$. 

\begin{table*}[ht]
   \centering
   \small
   \setlength{\extrarowheight}{1.2pt}
     \begin{tabular}{|c|c|cccccccc|c|}
     \hline

           & & \multicolumn{8}{c}{Dataset / Prompts}                       &  \\
           \cline{1-11}
     ID & Model & 1     & 2     & 3     & 4     & 5     & 6     & 7     & 8     & Average \\
     \hline
    
     1 & RNN (Last) & 0.524 & 0.025 & 0.004 & 0.001 & 0.001 & 0.004 & 0.165 & 0.094 & 0.102 \\
     2 & GRU (Last) & 0.521 & 0.265 & 0.274 & 0.678 & 0.441 & 0.563 & 0.420 & 0.182 & 0.418 \\
     3 & LSTM (Last) & 0.319 & 0.200   & 0.317 & 0.690  & 0.389 & 0.522 & 0.423 & 0.189 & 0.467 \\

     4 & RNN (Mean) & 0.597  &  0.488 &  0.603 &  0.745 &  0.740 &   0.759 &  0.741  & 0.489  & 0.645 \\
     5 & GRU (Mean) & 0.608 & 0.515 & 0.593 & 0.737 & 0.725 & 0.738 & 0.733 & 0.515 & 0.646 \\
     6 & LSTM (Mean) & 0.583 & 0.523 & 0.591 & 0.757 & 0.737 & 0.756 & 0.706 & 0.514 & 0.646 \\

     7 & BI-LSTM (Mean) & 0.794  & 0.625  & 0.665 &  0.674 &  0.776 &  0.613 &  0.679 &  0.506 &  0.667   \\
     8 & \textsc{SkipFlow} CNN (Bilinear) & 0.780  &  0.620 &   0.628 &  0.719  & 0.775 &  0.721 &  0.729  & 0.409 &  0.673 \\
     9 &  ML-LSTM (L=2) & 0.800  &  0.630 &   0.667  & 0.687 &   0.774 &  0.612 &  0.728 &  0.545  & 0.676 \\
    10 &  ML-LSTM (L=3) & 0.700  &  0.554  & 0.641  & 0.753  &  0.780 &   0.764 &   0.752 &  0.558 &  0.688  \\

     11 &  \textsc{SkipFlow} CNN (Tensor) &  0.782  &  0.657 &   0.666 &   0.727 &   0.781  &  0.756  &  0.757 &   0.440  &   0.696 \\
     12 & EASE$^{\star}$ (SVR) & 0.781 & 0.630  & 0.621 & 0.749 & 0.782 & 0.771 & 0.727 & 0.534 & 0.699 \\
     13 & EASE$^{\star}$ (BLRR) & 0.761 & 0.621 & 0.606 & 0.742 & 0.784 & 0.775 & 0.730  & 0.617 & 0.705 \\

         14 & RNN$^{\dagger}$ ($d$=300)  & 0.793  & 0.667 &0.591  &0.752 & 0.713 & 0.770  & 0.784 & 0.576 & 0.706 \\
         15 & CNN & 0.789    & 0.674  & 0.590   & 0.742  & 0.726  & 0.757  & 0.771  &  0.614  & 0.708 \\ 

     16 & GRU$^{\dagger}$ ($d$=300)  & 0.792 & 0.666 & 0.592 & 0.757   & 0.727 &  0.753 & 0.779 & 0.649 & 0.714 \\
     17 & LSTM$^{\dagger, \phi}$ ($d$=300) & 0.792 & 0.676 & 0.583 & 0.745 & 0.725 & 0.765 & 0.780 & 0.651 & 0.715 \\
      18 &  ATT-LSTM  & 0.793  &  0.660  &  0.664  & 0.768  & 0.800 & 0.789 &   0.788  & 0.637 &  0.737    \\
     19 & ML-LSTM (L=4) & 0.793 &  0.682  & 0.643  &  0.769 &  0.777  & 0.772 &  0.789 &  \underline{0.683}  & 0.739 \\
     \hline
     20 & \textsc{SkipFlow} LSTM* (Bilinear) & \underline{0.830}  & \underline{0.678} & \underline{0.677} & \underline{0.778} & \underline{0.795} & \underline{0.807} & \underline{0.790}  & 0.670  & \underline{0.753} \\

     21 & \textsc{SkipFlow} LSTM* (Tensor) & \textbf{0.832} & \textbf{0.684} &    \textbf{0.695} &    \textbf{0.788} &    \textbf{0.815} &    \textbf{0.810} &    \textbf{0.800} &    \textbf{0.697} &    \textbf{0.764} \\
     \hline
     \end{tabular}%
     \caption{Experimental results of all compared models on the ASAP dataset. Best result is in bold and 2nd best is underlined. Results are sorted by average performance. $\dagger$ denotes our implementation of a model from \cite{DBLP:conf/emnlp/TaghipourN16}, $\phi$ denotes the baseline for statistical significance testing, $*$ denotes statistically significant improvement. $\star$ denotes non deep learning baselines. }
   \label{tab:single_mdls}%
 \end{table*}%

\subsection{Baselines and Implementation Details}

In this section, we discuss the competitor algorithms that are used as baselines for our model. 

\begin{itemize}

\item \textbf{EASE} - The major non deep learning system that we compare against is the \textit{Enhanced AI Scoring Engine} (EASE). This system is publicly available, open-source\footnote{http://github.com/edx/ease} and also took part in the ASAP competition and ranked third amongst 154 participants. EASE uses manual feature engineering and applies different regression techniques over the handcrafted features. Examples of the features of EASE include length-based features, POS tags and word overlap. We report the results of EASE with the settings of Support Vector Regression (SVR) and Bayesian Linear Ridge Regression (BLRR).

\item \textbf{CNN} - We implemented a CNN model using $1D$ convolutions similar to \cite{DBLP:conf/emnlp/TaghipourN16}. We use a filter width of $3$ and a final embedding dimension of $50$. The outputs from the CNN model are passed through a mean pooling layer and finally through the final linear layer. 

\item \textbf{RNN / GRU / LSTM} - Similar to \cite{DBLP:conf/emnlp/TaghipourN16}, we implemented and tested all RNN variants, namely the vanilla RNN, GRU (Gated Recurrent Unit) and LSTM. We compare mainly on two settings of \textit{mean pooling} and \textit{last}. In the former, the average vector of all outputs from the model is used. In the latter, only the last vector is used for prediction. A fully connected linear layer connects this feature vector to the final sigmoid activation function. We use a dimension of $50$ for all RNN/GRU/LSTM models. 

\item \textbf{LSTM Variants} - Additionally, we also compare with multiple LSTM variants such as the Attention Mechanism (ATT-LSTM), Bidirectional LSTM (BI-LSTM) and the Multi-Layer LSTM (ML-LSTM). We use the AttentionCellWrapper implementation in TensorFlow with an attention width of $10$.

\end{itemize}
\subsubsection{Our Models} 

We compare two settings of our model, namely the bilinear and tensor composition. They are denoted as \textsc{SkipFlow} LSTM (Bilinear) and \textsc{SkipFlow} LSTM (Tensor) respectively. The bilinear setting is formally described as $s(a,b) = a^{T}\: \textbf{M} \:b$, where $a,b$ are vectors of two distant LSTM outputs and $\textbf{M}$ is a similarity matrix. The bilinear setting produces a scalar value, similar to the output of the tensor layer. The tensor layer, aside from the combination of multiple bilinear products, also includes a separate linear layer along with a non-linear activation function. For the tensor setting, the number of slices of the tensor is tuned amongst $\{2,4,6,8\}$. For both models, the hidden layer is set to $50$. There is \textbf{no} dropout for this layer and the bias vector is set to $0$. The relevance width of our model $\delta$ is set amongst $\{20,50,100\}$. In addition, to demonstrate the effectiveness and suitability of the LSTM model for joint modeling of semantic relevance, we conduct further experiments with the \textsc{SkipFlow} extension of the CNN model which we call the \textsc{SkipFlow} CNN. Similarly, we apply the same procedure on the convolved representations. Aside from swapping the LSTM for a CNN, the entire architecture remains identical.

To facilitate fair comparison, we implemented and evaluated \textbf{all} deep learning models ourselves in TensorFlow. We also implemented the architectures of \cite{DBLP:conf/emnlp/TaghipourN16} which we denote with $\dagger$. For training, the ADAM optimizer \cite{DBLP:journals/corr/KingmaB14} was adopted with a learning rate amongst $\{0.01,0.001,0.0001\}$ and mini-batch size amongst $\{64,128,256\}$. The gradient of the norm is clipped to $1.0$. The sequences are all padded with zero vectors up till the total maximum length\footnote{We used the dynamic RNN in TensorFlow in our implementation.}. We use the same embeddings from \cite{DBLP:conf/emnlp/TaghipourN16} and set them to trainable parameters. All experiments are conducted on a Linux machine running two GTX1060 GPUs.

\subsection{Experimental Results}
Table \ref{tab:single_mdls} reports the empirical results of all deep learning models. First, it is clear that the mean pooling is significantly more effective as compared to the last LSTM output. In the \textit{last} setting, the performance of RNN is significantly worse compared to LSTM and GRU possibly due to the weaker memorization ability. However, the performance of LSTM, GRU and RNN are similar using the \textit{mean pooling} setting.  This is indeed reasonable because the adoption of a mean pooling layer reduces the dependency of the model's memorization ability due to implicit access to all intermediate states. Overall, we observe that the performance of LSTM and GRU is quite similar with either \textit{mean pooling} or \textit{last} setting. Finally, we note that the performance of CNN is considerably better than RNN-based models. We also observe that a multi-layered LSTM performs considerably better than a single-layered LSTM. We also observe that adding layers also increases the performance. On the other hand, the bidirectional LSTM did not yield any significant improvements in performance. The performance of ATT-LSTM is notably much higher than the base LSTM. The best performing LSTM model is a multi-layered LSTM with $4$ layers. 

Additionally, we observe that \textsc{SkipFlow} LSTM (Tensor) outperforms the baseline LSTM (Mean) by almost $10\%$ in QWK score. Evidently, we see the effectiveness of our proposed approach. The tensor setting of \textsc{SkipFlow} LSTM is considerably better than the bilinear setting which could be due to the richer modeling capability of the tensor layer. On the other hand, we also note that the \textsc{SkipFlow} extension of CNN model did not increase the performance of CNN. As such, we see that the \textsc{SkipFlow} mechanism seems to only apply to the compositional representations of recurrent-based models. Moreover, the width of the CNN is $3$ which might be insufficient to offset the impreciseness of our fixed width matching. 

Finally, we compare \textsc{SkipFlow} LSTM with deep learning models\footnote{For fair comparison, we only compare against single neural models and not against ensemble approaches.} of \cite{DBLP:conf/emnlp/TaghipourN16}. The key difference is that these models (denoted with $\dagger$ in Table \ref{tab:single_mdls}) have a higher dimensionality of $d=300$. First, we observe that a higher dimensionality improves performance over $d=50$. Our \textsc{SkipFlow} LSTM (Tensor) outperforms LSTM$^{\dagger} (d=300)$ significantly by $5\%$. The performance of LSTM$^{\dagger} (d=300)$ and GRU$^{\dagger} (d=300)$ are in fact identical and are only slightly better than feature engineering baselines such as EASE (BLRR). We also observe that ATT-LSTM and ML-LSTM (L=4) with both $d=50$ also consistently outperform LSTM$^{\dagger}$ and GRU$^{\dagger}$. Conversely, our \textsc{SkipFlow} LSTM (Tensor) model outperforms the best feature engineering baseline (EASE) by about $6\%$. 

\subsection{Comparison against Published Results}
Finally we compare with published state-of-the-art results from \cite{DBLP:conf/emnlp/TaghipourN16}. While our reproduction of the vanilla LSTM ($d=300) $could not achieve similar results, our \textsc{SkipFlow} model still outperforms the reported results with a much smaller parameterization ($d=50$ instead of $d=300$). \textsc{SkipFlow} also remains competitive to an ensemble of 20 models (CNN + LSTM) with just a single model. 

\begin{table}[htbp]
  \centering
  \small
    \begin{tabular}{|c|c|}
    \hline
    System &  QWK \\
    \hline
    
    LSTM w/o MOT & 0.540 \\
    LSTM + Attention & 0.731 \\
    CNN + LSTM &0.708 \\
    BiLSTM &  0.699 \\
    LSTM (d=300) & 0.746 \\
     10 x CNN ensemble  & 0.726 \\
    10 x LSTM ensemble   & 0.756 \\
    20 x LSTM + CNN Ensemble & 0.761 \\
    \hline
    \textsc{SkipFlow} LSTM (Tensor) &   \textbf{0.764} \\
   \hline
    
    \end{tabular}%
    \caption{Comparision against published works. Single model of \textsc{SkipFlow} outperforms model ensembles.}
    \label{tab:pub_compare}
    \end{table}
\subsection{Runtime and Memory}
Table \ref{tab:runtime} reports the runtime and parameters of several LSTM variants. We observe that the runtime of our models only incur a small cost of 1-2 seconds over the baseline LSTM model. Our model also not only outperforms LSTM$^{\dagger}$ and ML-LSTM (L=4) in terms of QWK score but also in terms of memory footprint and runtime. \textsc{SkipFlow} is also faster then the attention mechanism (ATT-LSTM).
\begin{table}[htbp]
  \centering
\small
    \begin{tabular}{|c|c|c|}
    \hline
    Model & Epoch/s & \# Param \\
    \hline
    LSTM  &  8s     &  13K\\
        LSTM$^{\dagger}$ (d=300)  & 12s     & 450K \\
        BI-LSTM  & 18s & 13K \\
        ML-LSTM (L=4)  & 27s & 50K \\
        ATT-LSTM & 20s & 15K  \\
         \textsc{SkipFlow} LSTM (Bilinear) & 9s       &  18K\\
        \textsc{SkipFlow} LSTM (Tensor)  & 10s      & 25K \\
        \hline
    \end{tabular}%
    \caption{Comparisons of runtime and parameter size on prompt 1. All models are $d=50$ unless stated otherwise.}
  \label{tab:runtime}%
\end{table}%

\section{Effect of Hyperparameters}
In this section, we investigate the effect of hyperparameters, namely the number of tensor slices $k$ and the relevance width $\delta$. While we report the results on the test set, it is good to note that the curves on the development set follow exactly the same rank and pattern. 

\subsection{Effect of Tensor Slices on Performance}

\begin{figure}[ht]
  
  \centering
    \includegraphics[totalheight=3.2cm,width=0.48\textwidth]{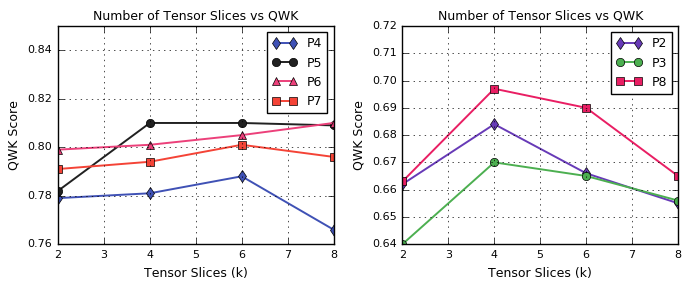}
    \caption{Effect of tensor slices on performance with $\delta=100$.}
    \label{fig:tensor_effect}
\end{figure}

Figure \ref{fig:tensor_effect} shows the effect of the number of tensor slices ($k$) on performance. The prompts\footnote{We omit prompt 1 because it has a much higher score which distorts the visualization of our graphs.} are separated into two graphs due to the different ranges of results. The optimal $k$ value is around $4$ to $6$ across all prompts. Intuitively, a small $k$ (2) and an overly large $k$ (8) often result in bad performance. The exception lies in prompts 5 and 6 where increasing the number of slices to $k=8$ either improved or maintained the QWK score. 

\subsection{Effect of Relevance Width $\delta$ on Performance}

\begin{figure}[ht]
  
  \centering
    \includegraphics[totalheight=3.2cm,width=0.48\textwidth]{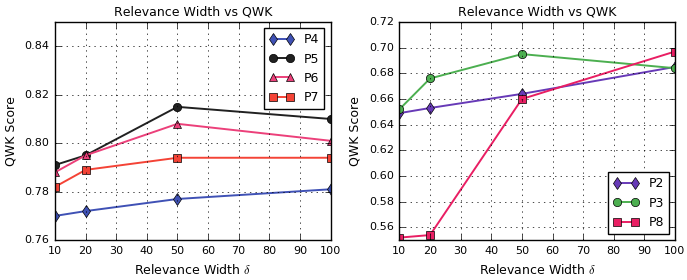}
    \caption{Effect of relevance width $\delta$ on performance with tensor slices $k=4$.}
    \label{fig:width_effect}
\end{figure}

Figure \ref{fig:width_effect} shows the influence of the hyperparameter relevance width $\delta$ on the performance. We observe that a small width produces worse results as compared to a large width. This is possibly due to insufficient tensor parameters or underfitting in lieu of a large number of matches is required with a small width. For example, consider prompt 8 that has the longest essays. Adopting $\delta=20$ for prompt 8 requires about $\approx 300$ to $400$ comparisons that have to be modeled by a fixed number of tensor parameters. A quick solution is to increase the size of the tensor. However, raising both $\delta$ and $k$ would severely increase computational costs. Hence, a trade-off has to be made between $\delta$ and $k$. Empirical results show that a value from 50 to 100 for $\delta$ works best with $k=4$. 

\section{Conclusion}
In this paper, we proposed a new deep learning model for Automatic Text Scoring (ATS). We incorporated the intuition of textual coherence in neural ATS systems. Our model, \textsc{SkipFlow} LSTM, adopts parameterized tensor compositions to model the relationships between different points within an essay, generating neural coherence features that can support predictions. Our approach outperforms a baseline LSTM on the same setting by approximately $10\%$ and also produces significantly better results as compared to multi-layered and attentional LSTMs. In addition, we also achieve a significant $6\%$ improvement over feature engineering baselines. 

\section{Acknowledgements}
The authors would like to thank anonymous reviewers of AAAI 2018, EMNLP 2017 and ACL 2017 whom have helped improve this work. 
\bibliography{references} 
\bibliographystyle{aaai}
\end{document}